\documentclass[letterpaper]{article} % DO NOT CHANGE THIS
\usepackage{aaai24}  % DO NOT CHANGE THIS
\usepackage{times}  % DO NOT CHANGE THIS
\usepackage{helvet}  % DO NOT CHANGE THIS
\usepackage{courier}  % DO NOT CHANGE THIS
\usepackage[hyphens]{url}  % DO NOT CHANGE THIS
\usepackage{graphicx} % DO NOT CHANGE THIS
\usepackage{natbib}  % DO NOT CHANGE THIS AND DO NOT ADD ANY OPTIONS TO IT
\usepackage{caption} % DO NOT CHANGE THIS AND DO NOT ADD ANY OPTIONS TO IT
\usepackage{algorithm}
\usepackage{algorithmic}

\usepackage{subfigure}
\usepackage{multirow}
\usepackage{makecell}
\usepackage{booktabs}
\usepackage{amsmath}
\usepackage{bm}
\usepackage{etoolbox}
\usepackage{amssymb}

\usepackage{bbding}
\usepackage{newfloat}
\usepackage{listings}
\DeclareCaptionStyle{ruled}{labelfont=normalfont,labelsep=colon,strut=off} % DO NOT CHANGE THIS
\floatstyle{ruled}
\newfloat{listing}{tb}{lst}{}
\floatname{listing}{Listing}
\title{Improving Factual Error Correction by Learning to Inject Factual Errors}
\author{
    Xingwei He\textsuperscript{\rm 1},
    Qianru Zhang\textsuperscript{\rm 1},
    A-Long Jin\textsuperscript{\rm 1},
    Jun Ma\textsuperscript{\rm 1},
    Yuan Yuan\textsuperscript{\rm 2,3,4}\thanks{Corresponding authors.},
    Siu Ming Yiu\textsuperscript{\rm 1}\footnotemark[1]
}
\affiliations{
    \textsuperscript{\rm 1}The University of Hong Kong, Hong Kong, China\\
    \textsuperscript{\rm 2}School of Computer Science and Engineering, Beihang University, Beijing, China\\
    \textsuperscript{\rm 3}State Key Laboratory of Software, Development Environment, 
    \textsuperscript{\rm 4}Zhongguancun Laboratory\\
    
    hexingwei15@gmail.com,
    qrzhang@cs.hku.hk,
    ajin@eee.hku.hk,\\
    junma@hku.hk,
    yuan21@buaa.edu.cn,
    smyiu@cs.hku.hk
}

\usepackage{bibentry}
\begin{document}

\maketitle

\begin{abstract}
Factual error correction (FEC) aims to revise factual errors in false claims with minimal editing, making them faithful to the provided evidence. This task is crucial for alleviating the hallucination problem encountered by large language models. 
Given the lack of paired data (i.e., false claims and their corresponding correct claims), existing methods typically adopt the `\textit{mask-then-correct}' paradigm. This paradigm relies solely on unpaired false claims and correct claims, thus being referred to as distantly supervised methods. 
These methods require a masker to explicitly identify factual errors within false claims before revising with a corrector. 
However, the absence of paired data to train the masker makes accurately pinpointing factual errors within claims challenging. 
To mitigate this, we propose to improve FEC by Learning to Inject Factual Errors (LIFE), a three-step distantly supervised method: `\textit{mask-corrupt-correct}'.  
Specifically, we first train a corruptor using the `\textit{mask-then-corrupt}' procedure, allowing it to deliberately introduce factual errors into correct text. The corruptor is then applied to correct claims, generating a substantial amount of paired data.  
After that, we filter out low-quality data, and use the remaining data to train a corrector. 
Notably, our corrector does not require a masker, thus circumventing the bottleneck associated with explicit factual error identification. 
Our experiments on a public dataset verify the effectiveness of LIFE in two key aspects: Firstly, it outperforms the previous best-performing distantly supervised method by a notable margin of 10.59 points in SARI Final (19.3\% improvement).  
Secondly, even compared to ChatGPT prompted with in-context examples, LIFE achieves a superiority of 7.16 points in SARI Final.
% Secondly, it even surpasses ChatGPT prompted with in-context examples by 7.16 points in SARI Final. 

\end{abstract}

\section{Introduction}

\begin{figure}
  \centering
    \includegraphics[width=0.47\textwidth]{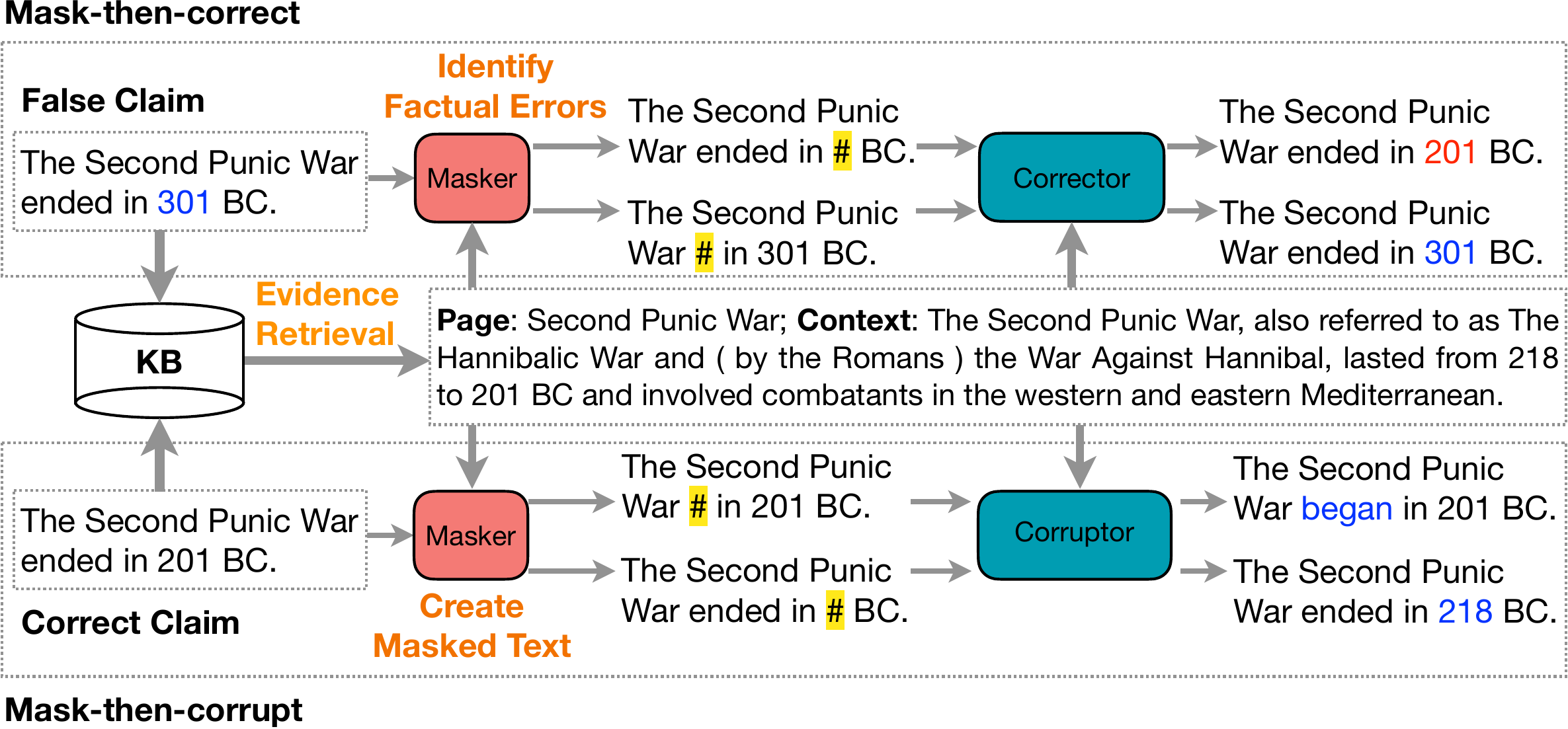} 
      \caption{ Comparison between the `\textit{mask-then-correct}' and `\textit{mask-then-corrupt}' pipelines during inference. 
      Text in blue and red denotes the factual errors and correct revision, respectively. \# refers to the mask token.
      % Text highlighted in blue and red indicates factual errors and their corresponding corrected revisions.
      }\label{fig.comparison}
\end{figure}

As is well known, large language models (LLMs), such as GPT-3 \cite{NEURIPS2020_1457c0d6}, PaLM\cite{chowdhery2022palm}, and LLaMA \cite{touvron2023llama}, have revolutionized NLP, which possess an extensive number of parameters and undergo pre-training on vast amounts of data. 
% The largest GPT-3 model, for instance, boasts 173 billion parameters. 
Compared with small language models, LLMs present emergent abilities \cite{wei2022emergent}, including in-context learning and reasoning \cite{wei2022chain} capabilities. 
% As a representative of LLMs, the recent launched ChatGPT has attracted widespread global attention, since it can produce coherent and contextually appropriate responses in various conversational contexts. 
A representative of LLMs is the recently launched ChatGPT, which has attracted widespread  attention due to its capacity to generate coherent and contextually appropriate responses across various conversational contexts. 
However, LLMs are prone to hallucinate unintended text, namely generating unfaithful, nonsensical, or factually incorrect text, a phenomenon referred to as `\textit{hallucination}' \cite{maynez-etal-2020-faithfulness, raunak-etal-2021-curious}. 
\textit{Factual error correction} (FEC) is meant to correct factual errors within the text to make the output factually consistent with the input. 
This task is crucial for alleviating the hallucination problem, as FEC models can be employed to post-edit text generated by LLMs, thereby enhancing their reliability and faithfulness.

The most effective and straightforward approach is to develop fully supervised FEC models by fine-tuning pre-trained models, such as T5 \cite{10.5555/3455716.3455856} on parallel data, which consists of false claims and their corresponding correct claims. 
However, it is very labor-intensive and time-consuming to create parallel data for FEC with human annotation, thus limiting the availability of such paired data. 
Therefore, previous methods \cite{shah2020automatic, thorne-vlachos-2021-evidence, chen2023converge} have focused on developing FEC models in a distantly supervised manner, based on the assumption that unpaired false claims and correct claims are readily available. 
Distantly supervised models generally adhere to the `\textit{mask-then-correct}' paradigm. 
To be concrete, during training, a masker is used to mask a correct claim, following which a corrector is optimized to reconstruct the original correct claim.
While during testing, the masker is tasked with identifying factual errors within a false claim, and then the corrector is expected to generate a correct claim based on the masked claim. 
For instance, as shown in the top of Figure \ref{fig.comparison}, during testing, the masker model is expected to precisely identify the factual error (i.e., `\textit{301}') in the false claim ``\textit{The Second Punic War ended in 301 BC}'', and then the corrector will generate the correct claim based on the masked text ``\textit{The Second Punic War ended in \# BC}''. 
However, since there is a lack of paired data to train the masker, it is non-trivial to accurately identity the factual error in the false claim. 
If the masker misidentifies the factual error (e.g., `\textit{ended}') in the mentioned false claim, it will lead to unreasonable masked text. 
As a result, the corrector is unlikely to generate the correct claim.

To circumvent the bottleneck of identifying factual errors before making corrections, we propose to improve factual error correction by \textbf{L}earning to \textbf{I}nject \textbf{F}actual \textbf{E}rrors into correct claims, referred to as \textbf{LIFE}. 
LIFE is a distantly supervised model, leveraging unpaired false claims and correct claims, yet it resorts to a three-step pipeline called `\textit{mask-corrupt-correct}'. 
The main motivation behind LIFE is to train a corruptor to inject factual errors into correct text. 
Suppose we have an ideal corruptor, we first use a masker to create diverse masked text, and then the corruptor will generate false claims based on these masked claims. 
For example, in the bottom of Figure \ref{fig.comparison}, the corruptor injects factual errors by replacing `\textit{ended}' with `\textit{began}' or substituting `\textit{201}' with `\textit{218}'. 
Consequently, we can create a sufficient collection of paired wrong claims and correct claims, which are utilized for training the corrector. 
% During testing, the masker in our proposed `\textit{mask-then-corrupt}' pipeline does not need to take on the responsibility of identifying factual errors, compared with the masker in previous `\textit{mask-then-correct}' pipeline, thus avoiding the above mentioned bottleneck. 
It is worth mentioning that our proposed `\textit{mask-then-corrupt}' pipeline eliminates the need for the masker to identify factual errors during testing, unlike the masker in the previous `\textit{mask-then-correct}' pipeline, thus bypassing the aforementioned bottleneck.

So far, the critical challenge lies in training an effective corruptor capable of introducing factual errors into correct claims. To achieve this, the masker masks certain words within a false claim, prompting the corruptor to reconstruct the original false claim based on the masked claim. 
In this way, the corruptor progressively grasps the skill of fabricating false claims by intentionally injecting factual errors into correct claims.

% To summarize, the main contributions of this work are as follows:
To summarize, our contributions are threefold: 
(1) We propose \textbf{LIFE}\footnote{Our code is available at: \url{https://github.com/
NLPCode/LIFE}.}, a distantly supervised model driven by the innovative three-step strategy: `\textit{mask-corrupt-correct}'. 
(2) During testing, LIFE can revise false claims in an end-to-end manner, eliminating the need for the masker to identify factual errors before correction. Therefore, the proposed model nicely bypasses the bottleneck encountered by previous distantly supervised methods. 
(3) Our experimental results on a public dataset demonstrate the superior performance of LIFE compared to previous distantly supervised baselines and few-shot LLMs. LIFE achieves a remarkable state-of-the-art (SOTA) result on the test set, scoring 65.59 on SARI Final and 66.51 on ROUGE-2. These compelling outcomes validate the effectiveness of our proposed approach.

\section{Approach}

\subsection{Problem Statement}
Factual error correction aims to rectify factual errors within claim $C$ using minimal revisions, making the revised claim $C^\prime$ align with the provided evidence $E$. 
Factual error correction follows three requirements: 
The revised claim $C^\prime$ should be grammatical, supported by the evidence, and rectify the factual errors present in $C$. 

\subsection{Overview}
Following previous distantly supervised methods, we assume that unpaired false claims and correct claims are available.
The set of false claims and correct claims are referred to as $\mathcal{S}^f=\{C^{f}_1, \dots, C^{f}_n\}$ and 
$\mathcal{S}^t=\{C^{t}_1, \dots, C^{t}_m\}$, where $C^{f}_i$ and  $C^{t}_i$ denote the $i$-th false and correct claims within their respective sets. $C^{f}_i$ and $C^{t}_i$ are unpaired, namely $C^{t}_i$ is not the correct claim for $C^{f}_i$. 
In this work, we propose LIFE to bypass the bottleneck of identifying factual errors before correction during testing. Our propose model consists of three key modules: a masker, a corruptor and a corrector. 
% The masker and corruptor are unified with the  `\textit{mask-then-corrupt}' pipeline. 
The masker and corruptor work in tandem through a `\textit{mask-then-corrupt}' pipeline.
Upon being trained on $\mathcal{S}^f$, the corruptor is developed, enabling us to subsequently introduce factual errors into each correct claim from $\mathcal{S}^t$. 
In this way, we can create a substantial amount of synthetic data, which will be used to train the corrector.
% After training on $\mathcal{S}^f$, we will obtain a corruptor, and then we will use it inject factual errors into each correct claims in $\mathcal{S}^t$. In this way, we can create enough synthetic data, which will be used to train the corrector.

In the following subsections, we delve into the training process of the corruptor and its application in generating synthetic data during testing. 
Finally, we will introduce the utilization of filters to refine the synthetic data, followed by the training of the corrector using these refined data. 
% Finally, we will introduce how to use filters to refine the synthetic data, and then train the corrector on them. 

\begin{figure*}
  \centering
    \subfigure[The `\textit{mask-then-corrupt}' pipeline]{
      \centering
      \includegraphics[width=0.62\textwidth]{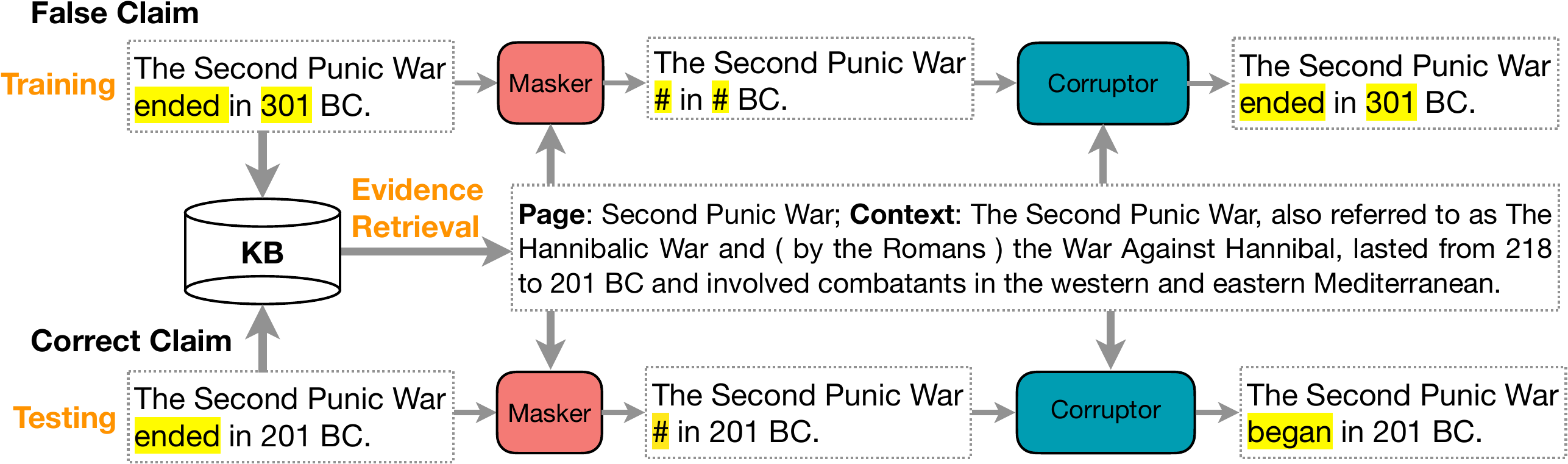} 
    }
    % \hspace{-1mm}
    \subfigure[Illustration of synthetic data filtering.]{
      \centering
      \includegraphics[width=0.29\textwidth]{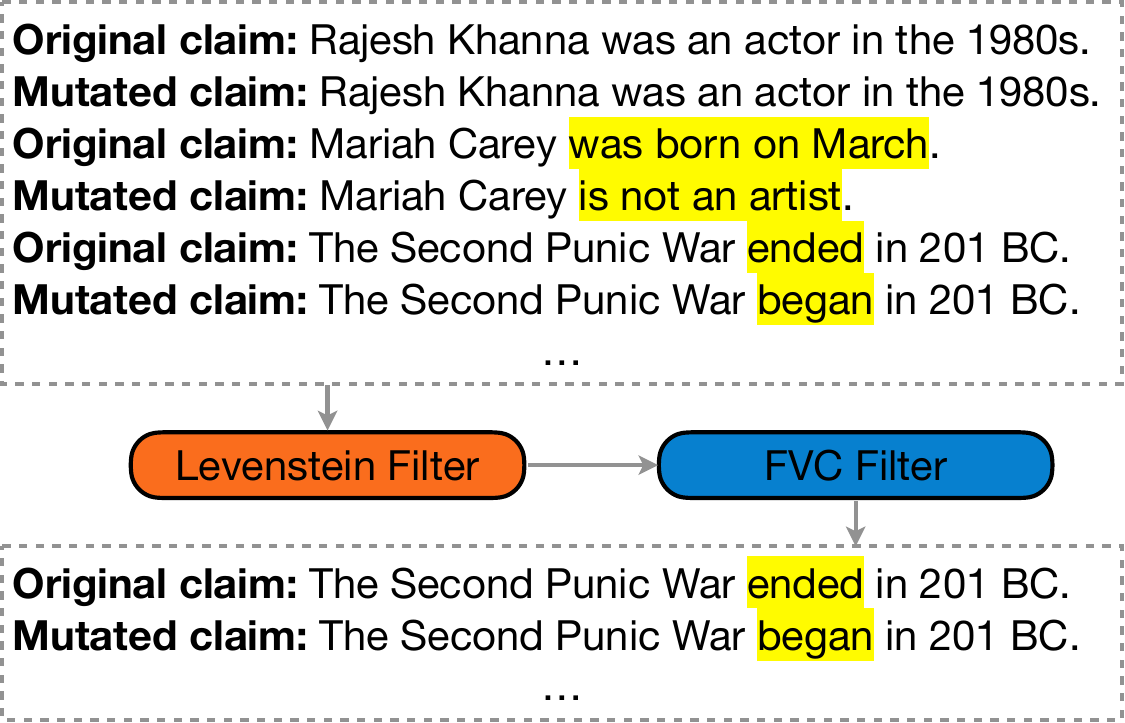} 
    }
      \caption{ The corruptor is trained to reconstruct the false claim, conditioned on the masked claim and retrieved evidence. At test time, the corruptor is able to incorporate relevant factual errors into the correct claim to generate a false claim. \# denotes the mask token. The masked words and newly predicted words are highlighted.  
      }
      \label{fig.corruptor_filter}
\end{figure*}

\subsection{Corruptor Training and Inference}\label{sec.corruptor}
As depicted in Figure \ref{fig.corruptor_filter} (a), the  `\textit{mask-then-corrupt}' pipeline comprises two essential components: a masker and a corruptor. In the following, we will introduce two kinds of maskers, and how to effectively train and test the corruptor.

\paragraph{Masker.}
Following \citet{thorne-vlachos-2021-evidence}, we resort to two simple maskers to mask the input claims, \textbf{random masking} and \textbf{heuristic masking}. Random masking randomly masks some words within the input claim.  
On the other hand, heuristic masking masks words that appear in the claim but do not appear in the evidence. 

\paragraph{Corruptor.} The corruptor is an encoder-decoder transformer \cite{transformer}, designed to reconstruct the claim based on the masked claim and the given evidence. 

\paragraph{Training and Testing.} 
During training, we input a false claim $C^{f}\in\mathcal{S}^f$ into the masker, and then optimize the corruptor to effectively recover $C^{f}$. 
When the masker successfully identifies the factual errors within the false claim, it implies that the corruptor must introduce factual errors into the masked claim to restore $C^{f}$ accurately. 
In this scenario, the corruptor should deliberately introduce factual errors into the masked claim. However, if the masker fails to recognize any factual errors within the false claim, the corruptor simply needs to restore the masked words in the masked claim without requiring to inject additional factual errors. 

During testing, we feed a correct claim $C^{t}_i\in\mathcal{S}^t$ into the masker and expect the corruptor to generate a false claim $C^{g}_i$ based on the masked claim rather than reconstructing $C^{t}_i$. 
Accordingly, we acquire a set of synthetic paired data denoted as $\mathcal{D^\prime}=\{(C^{t}_1, C^{g}_1), \dots, (C^{t}_m, C^{g}_m)\}$, which is employed for training the corrector. Here, $C^{t}_i$ and $C^{g}_i$ represent the $i$-th correct claim and generated claim, respectively.

However, in reality, we cannot guarantee that the corruptor will always inject factual errors into the masked claim. Furthermore, there is no guarantee that the  generated claim will have a high correlation with the original claim.
It is possible for the corruptor to deviate from the masked input and create an entirely new claim. 
% rather than simply disregarding the masked claim and creating an entirely new claim. 
Therefore, it is necessary for us to apply additional filters to the generated data $\mathcal{D^\prime}$.

\subsection{Synthetic Data Filtering}\label{sec.filter}
In Figure \ref{fig.corruptor_filter} (b), we first filter out data instances, where $C^{t}_i$ is completely the same with $C^{g}_i$, and then apply the Levenshtein filter and fact verification classifier-based filter to the synthetic data. The filtered data is referred to as $\mathcal{D}$. 
\paragraph{Levenshtein Filter.} 
The Levenshtein filter (LF) is based on the character-level Levenshtein edit distance \cite{levenshtein1966binary} between  $C^{t}_i$ and $C^{g}_i$: 
\begin{eqnarray}\label{eq:1}
  LF(C^{t}_i, C^{g}_i) =  \frac{Levenshtein(C^{t}_i, C^{g}_i)}{length(C^{t}_i)}.
\end{eqnarray}
If $LF(C^{t}_i, C^{g}_i)$ exceeds a certain threshold $t_{l}$, the corresponding data instance will be excluded.  
This filter relies on the assumption that a significant disparity in the Levenstein distance between $C^{t}_i$ and $C^{g}_i$ indicates a considerable semantic deviation of $C^{g}_i$ from $C^{t}_i$.

\paragraph{Fact Verification Classifier-based Filter.} 
The fact verification classifier-based filter (FVCF) relies upon the fact verification classifier, a 3-way classifier trained on FEVER.  This filter is designed to classify a given claim as SUPPORTED, REFUTED, or NOTENOUGHINFO based on whether the given claim is supported, refuted, or unable to be verified by the provided evidence. 
If the predicted probability of NOTENOUGHINFO for the generated claim $C^{g}_i$ surpasses a threshold value $t_{c}$, this indicates that $C^{g}_i$ diverges significantly from the original claim $C^{t}_i$ at the semantic level. 
This deviation is so substantial that $C^{g}_i$ becomes unverifiable by the given evidence.
Therefore, ($C^{g}_i$, $C^{t}_i$) will be discarded.

% The disparity is to such an extent that it becomes unverifiable by the initial set of evidence.

% \begin{figure}
%   \centering
%     \includegraphics[width=0.3\textwidth]{./figures/corrector-crop.pdf} 
%       \caption{ Input format of the corrector.
%       }\label{fig.corrector}
% \end{figure}
% \subsection{Corrector Training}\label{sec.corrector}

We fine-tune T5 on the synthetic data $\mathcal{D}$, which serves as the corrector. 
% As shown in Figure \ref{fig.corrector}, 
The corrector takes the mutated claim $C^{g}_i$ and evidence as input. 
During training, our objective is to minimize the cross-entropy loss between the revised claim  $\hat{C^{t}_i}$ and the original correct claim  $C^{t}_i$.

% \begin{eqnarray}\label{eq:1}
%   P(C^{t}_i|C^f_i, E_i; \theta) = \prod_{n=1}^{N}p(C^{t}_{i, n}|C^{t}_{i, j<n}, C^f_i, E_i;\theta), \nonumber
% \end{eqnarray}
% where $\theta$ represents the parameters of the corrector, $E_i$ denotes the corresponding evidence, and $C^{t}_{i, j<n}$ refers to the sub-sequence preceding $C^{t}_{i, n}$.

\section{Experiment}
\subsection{Experimental Setups}
\begin{table}[t]
   % \footnotesize
  \scriptsize
    \centering
      \begin{tabular}
        % {l|rrr|r}
      {
       m{0.12\textwidth}<{\raggedright}|
       m{0.08\textwidth}<{\raggedleft}
       m{0.08\textwidth}<{\raggedleft}
       m{0.08\textwidth}<{\raggedleft}
       }
      % \toprule
      \hline

      \textbf{Label / Split} & \textbf{Train} & \textbf{Valid} & \textbf{Test} \\
      % \midrule
      \hline
      SUPPORTED & 37,961 & 1,477 & 1,593 \\
      REFUTED  & 20,075 & 2,091 & 2,289\\
      \hline
      % \midrule
      Total & 58,036 & 3,568 & 3,882 \\ 
    \hline
    % \bottomrule
    \end{tabular}
    \caption{The count of data instances for each split and label within FECDATA.
    }\label{tab.data}
\end{table}

\paragraph{Dataset.}
We assess the performance of our model using the evidence-based FEC dataset (FECDATA), a manually created data proposed by \citet{thorne-vlachos-2021-evidence}. This dataset originates from the extensive fact verification dataset known as FEVER \cite{thorne-etal-2018-fever}. 
The claims present in the FEVER dataset are categorized into three classes: SUPPORTED, REFUTED, or NOTENOUGHINFO, depending on whether the claim is supported, contradicted, or cannot be verified by the provided evidence. 
FECDATA is curated by selecting instances belonging to the SUPPORTED and REFUTED categories. 
Notably, the REFUTED subset assesses models' ability to rectify factual errors within false claims. Conversely, the SUPPORTED subset evaluates their capacity to maintain correct claims.
Please refer to Table \ref{tab.data} for the basic statistics of FECDATA.

\paragraph{Evaluation Metrics.}
For automatic evaluation, we use SARI \cite{xu-etal-2016-optimizing} and ROUGE-2 \cite{lin-2004-rouge} metrics\footnote{The evaluation codes for SARI and ROUGE are available at: \url{https://huggingface.co/spaces/evaluate-metric/sari} and \url{https://huggingface.co/spaces/evaluate-metric/rouge}, respectively.}, which    
exhibit a strong positive correlation with human evaluation, especially SARI, according to \citet{thorne-vlachos-2021-evidence}'s findings.
We present the \textbf{Keep}, \textbf{Delete}, and \textbf{Add} scores of SARI, to assess the words in the revised claim (output) that are retained, removed, and introduced from the mutated claim (input), compared with the reference (ground truth). 
The SARI \textbf{Final} denotes the average of these three scores. ROUGE-2 (\textbf{RG-2}) measures the number of matching bigrams between the revised claim and the reference claim.

\paragraph{Baselines.} 
We compare our proposed model with three kinds of baselines: 
% fully supervised, distantly supervised and few-shot methods. 

\textbf{Fully Supervised Baselines} are used to assess the ceiling performance of FEC models, assuming the availability of manually created paired data for training. 
Following \citet{thorne-vlachos-2021-evidence}, we fine-tune \textbf{T5-base} \cite{10.5555/3455716.3455856} on the FECDATA training set, where the encoder takes the false claim and its corresponding evidence as input, while the decoder generates the revised claim.

\textbf{Distantly Supervised Baselines } employ the `\textit{mask-then-correct}' pipeline, which comprises a masker and a corrector.  
The masker can take various forms, such as the token-level explanations \cite{ribeiro-etal-2016-trust, chen-etal-2017-enhanced} of a fact verification classifier (FVC), random masking, or heuristic masking. 
The FVC is typically initialized with BERT \cite{devlin-etal-2019-bert} or RoBERTa  \cite{liu2019roberta} and is trained on FEVER. 
The last two maskers have been discussed in the approach section. 
The corrector is typically trained on the SUPPORTED data instances from FEVER. 
There are several specific approaches within this framework: 
(1) Dual encoder pointer network (\textbf{DEPN}) \cite{shah2020automatic} uses an FVC as the masker and employs the dual encoder pointer generator with the copy mechanism \cite{see-etal-2017-get} as the corrector. 
(2) T5 Masker-Corrector (\textbf{T5MC}) \cite{thorne-vlachos-2021-evidence} differs from DEPN in two aspects: (a) It utilizes random masking during training and heuristic masking during testing.  (b) The corrector is based on T5-base. 
(3) Unlike T5MC, \textbf{T5MC-MLM} uses the masked language model BERT as the masker during inference. 
(4) \textbf{T5MC-V} is a variant of T5MC, using an FVC as the masker. 
(5) \textbf{VENCE} \cite{chen2023converge} iteratively runs the `\textit{mask-then-correct}' pipeline over the claim until it becomes supported by evidence or the algorithm reaches the maximum steps.

\textbf{Few-shot Baselines} include two types of models: (1) \textbf{8-shot T5-base} directly fine-tunes T5-base using 8 data examples. 
(2) \textbf{8-shot LLMs} correct false claims via few-shot in-context learning, where we prompt three OpenAI LLMs: text-ada-001, text-babbage-001, 
% text-curie-001, 
and gpt-3.5-turbo-0301 (i.e., ChatGPT) with 8 in-context examples. Please refer to Table \ref{tab.fec.negate_few-shot} in Appendix  for the prompt applied to LLMs. 
For fair comparisons, all few-shot baselines use the same set of examples.

\subsection{Implementation Details}
% \section{Implementation Details}\label{sec.implementation}
\textbf{Evidence Retrieval. } 
As our work does not focus on evidence retrieval, we use the evidence retrieved by \citet{thorne-vlachos-2021-evidence} for all models. 
The retrieval process involves two steps: GENRE \cite{cao2021autoregressive} is first used to predict the relevant Wikipedia pages for the input claim; DPR \cite{karpukhin-etal-2020-dense} is then used to retrieve the top-$k$ ($k=2$) passages from the pages predicted by GENRE. \\
\textbf{Masker and Corruptor.}
For the masker, we employ a heuristic masking strategy to mask the provided claim during training, while random masking with a mask ratio of $30\%$ is utilized during inference. 
The corruptor is initialized with the T5-base model and 
optimized using the AdamW optimizer \cite{loshchilovdecoupled} with a learning rate of $4e-5$, a batch size of $64$, and a linear learning rate schedule with $10\%$ warm-up steps for $4000$ steps. The exploration of learning rates is conducted within the predefined set: $\{5e-6, 1e-5, 2e-5, 3e-5, 4e-5, 5e-5\}$. 
During inference, we use beam search decoding with a beam size of $5$ to generate false claims for all correct claims (i.e., SUPPORTED data instances) in the training and validation sets. 
The corruptor takes the top-$2$ retrieved/gold evidence paired with the masked claim as input. 
We set the maximum source length to 512 and the maximum target length to 256.\\
\textbf{Filters.} 
We initialize the fact verification classifier-based filter with RoBERTa-base and then fine-tune it on FEVER for $4000$ steps. 
We set the threshold $t_l=0.3$ and $t_c=0.2$ to filter the synthetic data produced by the corruptor. \\
\textbf{Corrector.} 
Following \citet{thorne-vlachos-2021-evidence}, the corrector is based on the T5-base model, which takes the top-$2$ retrieved/gold evidence paired with the input claim as input. 
% The corrector uses the same training and inference settings as the corruptor, with the exception of the training steps being set to 1000.
The corrector follows the corruptor's settings for training and inference, except it trains for only 1000 steps.

We employ the HuggingFace Transformers library \cite{Wolf2019HuggingFacesTS} to implement all models. Additionally, all experiments are carried out utilizing $2$ NVIDIA Tesla V100 GPUs, each equipped with 32 GB of memory.

\begin{table*}[t] 
  \centering
 % \footnotesize
  \scriptsize
   \begin{tabular}{
    m{0.28\textwidth}<{\raggedright}
    m{0.1\textwidth}<{\centering}
    m{0.03\textwidth}<{\centering}
    m{0.035\textwidth}<{\centering}
    m{0.03\textwidth}<{\centering}
    m{0.035\textwidth}<{\centering}
    m{0.03\textwidth}<{\centering}
    m{0.03\textwidth}<{\centering}
    m{0.035\textwidth}<{\centering}
    m{0.03\textwidth}<{\centering}
    m{0.035\textwidth}<{\centering}
    m{0.03\textwidth}<{\centering}
    }
    % \toprule
    \hline
    \multirow{3}{*}{\textbf{Models}} & \multirow{3}{*}{\textbf{FVC}}& \multicolumn{5}{c} {\textbf{Retrieved Evidence}} & \multicolumn{5}{c} {\textbf{Gold Evidence}} \\
    \cmidrule(lr){3-7} \cmidrule(lr){8-12}
     && \multicolumn{4}{c} {\textbf{SARI Score}} & \multirow{2}{*}{\textbf{RG-2}} & \multicolumn{4}{c} {\textbf{SARI Score }} & \multirow{2}{*}{\textbf{RG-2}}\\
    \cmidrule(lr){3-6} \cmidrule(lr){8-11}
    && \textbf{Keep} &  \textbf{Delete} & \textbf{Add} & \textbf{Final}  & & \textbf{Keep} &  \textbf{Delete} & \textbf{Add} & \textbf{Final} & \\
    % \midrule
    \hline
    \multicolumn{12}{l}{\textbf{Fully Supervised Baselines}} \\
    % Supervised BART-base$^{\ast}$  & -  & 70.75 & 65.65 & 38.88 & 58.43 &  59.99 & 73.51 & 69.27 & 47.30 & 63.36 & 64.00 \\
    Fully Supervised T5-base$^{\ast}$ & - &  85.40 & 88.92 & 48.40& \underline{74.24}& \underline{73.50} & 88.56 & 91.40 & 58.38 & \underline{79.45} & \underline{78.04} \\
    % \midrule
    \hline
    \multicolumn{12}{l}{\textbf{Distantly Supervised Baselines}} \\
    DEPN \cite{shah2020automatic}$^\ddagger$ & BERT-base & 34.5 & 48.1 & 1.7 & 28.1 & 34.8 & 45.2 & 56.9 & 3.9 & 35.3 &- \\
    T5MC \cite{thorne-vlachos-2021-evidence}$^\dagger$ & - & 65.2 & 62.7  & 15.5 & 47.8 & 50.3 & 66.7 & 62.2 & 16.1 & 48.3  & - \\
    \qquad  + Enumerate$^\dagger$ & BERT-base & 66.2 & 64.3 & 17.1 &  49.2 & 51.2  & - & -& -& -  &- \\
    T5MC-MLM$^\ddagger$ & - & 56.1 & 52.9 & 7.8 & 38.9 & 42.7 & - & -& -& -  &- \\
    T5MC-V \cite{thorne-vlachos-2021-evidence}$^\dagger$ & BERT-base & 61.1 & 54.3 & 19.4 & 44.9 & 42.0 & 61.8 & 62.2 & 10.2 & 44.7 & - \\
    \qquad  + Enumerate$^\dagger$ & BERT-base & 63.0 & 55.7 & 24.1 & 47.6& 45.5  & - & -& -& -  &- \\
    \multirow{2}{*}{VENCE \cite{chen2023converge}$^\dagger$} & BERT-base & 66.0 & 60.1 & 34.8 & 53.6 & 57.7 & 67.5 & 61.5 & 34.6 & 54.5 & - \\
     & RoBERTa-large & 67.1 & 61.9 & 36.0 & 55.0 & 59.1  & - & -& -& -  &- \\
    % \midrule
    \hline
    \multicolumn{12}{l}{\textbf{Few-shot Baselines}} \\
    8-shot T5-base$^{\ast}$ & - & 61.75& 85.23 & 8.70 & 51.89 & 49.83 &  63.50& 82.44& 13.56 & 53.17 & 51.38 \\
    8-shot text-ada-001$^{\ast}$ & - & 61.43 & 75.21 & 9.86 & 48.83 & 42.95 & - & -& -& -  &- \\
    8-shot text-babbage-001$^{\ast}$ & -  & 69.69 & 76.39 & 18.07 & 54.72 & 52.57 & - & -& -& -  &-\\
    % 8-shot text-curie-001$^{\ast}$ & -& 76.01 & 79.61 & 26.33 & 60.65 & 59.34 & - & -& -& -  &-\\
    8-shot ChatGPT$^{\ast}$ & -  & 72.09 & 75.92 & 27.29 & 58.43 & 49.43 &  79.98 & 81.61 & 38.81 & 66.80 & 60.72\\
    % \midrule
    \hline
    \multicolumn{12}{l}{\textbf{Distantly Supervised (Our Method)}} \\
    LIFE$^{\ast}$ &  RoBERTa-base & 75.23 & 91.88 & 29.67 & \textbf{65.59} & \textbf{66.51} & 79.33 &93.01 & 39.81 & \textbf{70.72} & \textbf{70.45} \\
    % \bottomrule
    \hline
 \end{tabular}
 \caption{Automatic evaluation results (\%) of different models with retrieved/gold evidence on the FECDATA test set. 
 % MLM does not use any evidence. 
 Results marked with $\dagger$, $\ddagger$, and $\ast$ are from VENCE, T5MC-V and our reproduction, respectively. 
% Bb and Rl denote BERT-base and RoBERTa-large. 
Enumerate means using the FVC model to rank 20 generated claims and select the best one. 
\underline{Underline} indicates the best model and \textbf{bold} indicates the second best. 
 % The number behind the model represents the size of the model's parameters.
 }\label{tab.main_result} 
\end{table*} 

\subsection{Overall Experimental Results}
We present the main experimental results on the FECDATA test set in Table \ref{tab.main_result}, which reveal the following key findings: \\
\textbf{Distantly supervised baselines based on the mask-then-correct approach perform worst.} This is mainly because these methods rely on the accurate identification of factual errors in the given claim by the masker. 
Excessive or incorrect masking of words in the claim by the masker will impair the performance of the corrector.\\
\textbf{LLMs demonstrate strong performance with a few examples.} 
Simply fine-tuning T5-base using 8 labeled instances (8-shot T5-base) cannot bring any enhancement compared to existing distantly supervised baselines, such as VENCE. 
By comparison, LLMs prompted with 8 in-context examples performs much better than 8-shot T5-base. 
For example, 8-shot ChatGPT achieves 58.43 points in SARI Final and surpasses VENCE by around 3 points, which  underscores the remarkable few-shot ability of LLMs for FEC. \\
% In comparison, ChatGPT, which is an LLM, can achieve a SARI Final score of 58.43 when presented with 8 in-context examples. This score is noteworthy and exceeds VENCE's score by approximately 3 points.
% By comparison, when LLMs such as ChatGPT are prompted with 8 in-context examples, they will achieve a notable SARI Final score of 58.43, surpassing VENCE by around 3 points.
\textbf{Our proposed LIFE achieves a new SOTA result.} 
LIFE outperforms its best distantly supervised counterpart, VENCE, by a significant margin of 10.59 points in the SARI Final score.
This notable achievement can be attributed to LIFE's effective circumvention of the bottleneck associated with mask-then-correct methods, which involves identifying factual errors in claims prior to the correction process. 
Notably, our approach even surpasses the performance of few-shot LLMs. These results offer compelling evidence for the effectiveness of LIFE in boosting the performance for FEC.

While LIFE outperforms distantly supervised and few-shot models, it still falls short of the fully supervised baseline (scoring 65.59 compared to 74.24 on SARI Final), highlighting the potential for further enhancement. 
In addition, we found that using gold evidence instead of retrieved evidence to modify the given claim has led to significant improvements across all models. This indicates that retrieved evidence may be inadequate, namely fails to provide useful information for rectifying factual errors within claims. 
Given that this work does not focus on enhancing retrieval models, this discovery unveils a promising direction for future advancements in FEC models.

% Furthermore, we found that using gold evidence instead of retrieved evidence to modify the given claim has led to significant improvements across all models. This indicates that retrieved evidence often fails to provide effective information. As our focus is not on enhancing retrieval models, this presents a breakthrough opportunity for future advancements in the FEC model.

\begin{figure}
  \centering
    \includegraphics[width=0.46\textwidth]{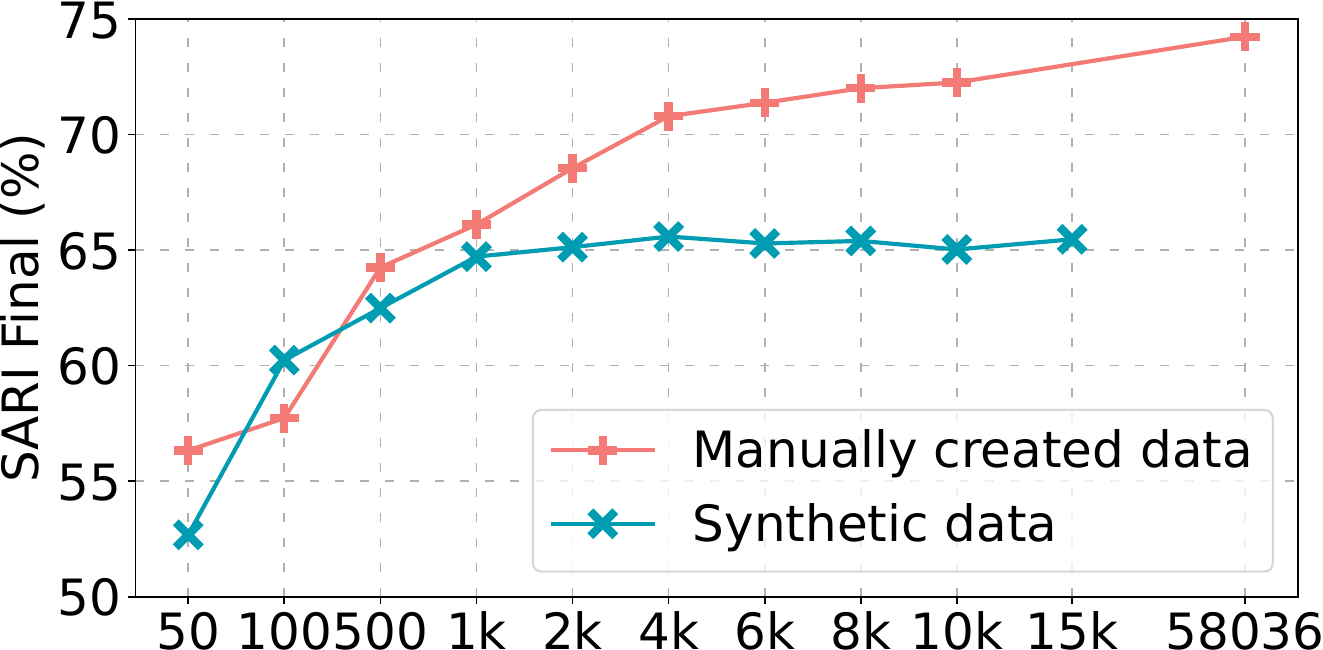} 
    \caption{ 
        Performance of correctors initialized with T5-base fine-tuned on different numbers of synthetic data or manually created data, i.e. FECDATA training data. 
      }\label{fig.effect}
\end{figure}

\subsection{More Analysis and Discussion}
In this subsection, unless explicitly stated otherwise, all experiments are conducted with the following settings: When training the corruptor, heuristic masking is used as the masker, while during testing, random masking is employed to generate synthetic data. Subsequently, the Levenshtein filter (LF) and the fact verification classifier-based filter (FVCF) are utilized to filter the generated data. The minimum masking granularity for the masker is at the word level, and adjacent masked words are not merged. Both the corrector and corruptor utilize the top-$2$ retrieved evidence.

\paragraph{How does the number of synthetic data affect the performance of LIFE?}\label{effect_synthetic} 
To answer the first question, we train LIFE's corrector using varying number of the synthetic data. 
For the sake of comparison, we also train a supervised corrector on the manually created FECDATA. 
As illustrated in Figure \ref{fig.effect}, when the number of data used to training correctors is less than $1k$, both correctors exhibit a linear increase in performance as the dataset size grows.
Notably, the corrector trained on synthetic data achieves a performance level that is comparable to the one trained on FECDATA. 
However, when the synthetic data reaches $4k$ samples, LIFE's performance reaches its peak. In contrast, the performance of the supervised corrector trained on gold data continues to improve, even when using the entire FECDATA training set. 
The significant performance gap between LIFE and the supervised corrector may stem from noise in synthetic data.

% The discernible performance disparity between LIFE and the supervised corrector trained on authentic FECDATA might potentially stem from the presence of noise within the synthetic data.

\begin{table}[t] 
  \centering
 % \footnotesize
  \scriptsize
   \begin{tabular}{
    m{0.07\textwidth}<{\raggedright}
    m{0.06\textwidth}<{\raggedright}
    m{0.03\textwidth}<{\centering}
    m{0.04\textwidth}<{\centering}
    m{0.03\textwidth}<{\centering}
    m{0.03\textwidth}<{\centering}
    m{0.03\textwidth}<{\centering}
    }
    % \toprule
     \hline
     \multicolumn{2}{c}{\textbf{Mask Strategy}} & \multicolumn{4}{c} {\textbf{SARI Score}} & \multirow{2}{*}{\textbf{RG-2}} \\
    \cmidrule(lr){3-6} 
    \cmidrule(lr){1-2} 
   \textbf{\#} \textbf{Train}&  \textbf{Test} & \textbf{Keep} &  \textbf{Delete} & \textbf{Add} & \textbf{Final}  & \\
   % \midrule
    \hline
  1 Random & Random & 70.20 & 93.61 & 16.31 & 60.04 & 63.15 \\
  2 Heuristic & Heuristic & 75.25 & 88.85 & 19.79 & 61.30 & 61.81\\
   % \midrule
    \hline
  3 Random & Heuristic & 72.09 & 86.68  & 16.85 & 58.54 & 60.03 \\
  4 Heuristic & Random & 75.23 & 91.88 & 29.67 & \textbf{65.59} & \textbf{66.51} \\
   % Heuristic & Random & 74.80 & 91.63 & 28.78  & \textbf{65.07} & \textbf{66.10}\\
    % \bottomrule
    \hline
 \end{tabular}
 \caption{Impact of masking strategies on LIFE. 
 }\label{tab.mask_stategy} 
\end{table} 

\begin{figure}
  \centering
    \includegraphics[width=0.46\textwidth]{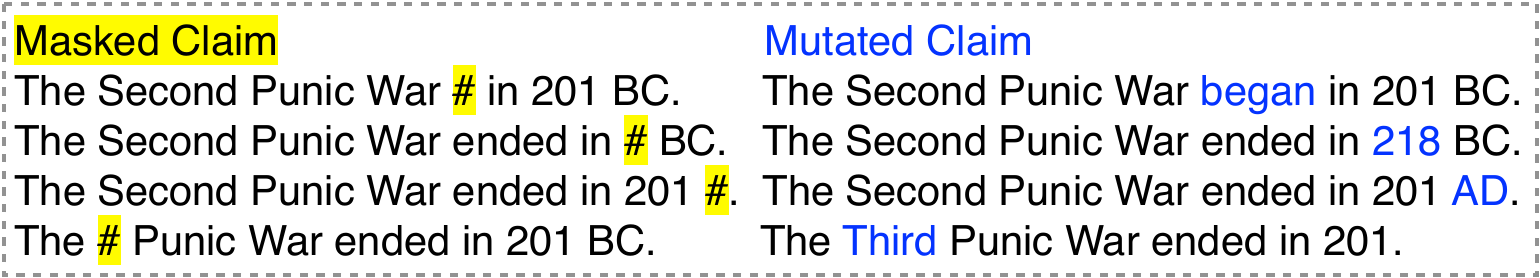} 
    \caption{ 
        Mutated claims generated based on varied masked claims. 
        The first masked claim is created by heuristic masking, the others are produced by random masking. Please refer to Figure \ref{fig.corruptor_filter} for the original correct claim and retrieved evidence 
         (\#=masked token, blue text=injected factual errors). 
      }\label{fig.generated_claim}
\end{figure}

\paragraph{How do different masking strategies affect the performance of LIFE?}
In this work, we employ two masking strategies to mask the given claim.  
We exhaustively enumerate all potential combinations of masking strategies utilized during both the training and testing phases of the LIFE corruptor. 
From Table \ref{tab.mask_stategy},  we draw two crucial findings: (1) 
The heuristic strategy, when employed during training, enhances the performance of LIFE more effectively than the random strategy (row 1 vs. row 4, and row 2 vs. row 3). 
(2) However, during testing, the random strategy proves to be more useful (row 1 vs. row 3, and row 2 vs. row 4). 

During training, a false claim is fed into the masker. The heuristic strategy is more likely to find  erroneous spans. 
For example, in Figure \ref{fig.corruptor_filter}, factual errors related words (\textit{`ended'} and \textit{`301'}) are masked due to their absence in the supporting evidence.  
Consequently, the corruptor is compelled to introduce factual errors to recover the original false claim.
By comparison, the random strategy is less likely to be able to accurately mask the problematic words. 
In such instances, the masked claim still contains factual errors, thereby the corruptor only needs to complete the masked claim without requiring to inject factual errors into it. At this point, the corruptor is not optimized in the expected direction. Hence, the heuristic strategy manifests greater efficacy during training. 

Conversely, during testing, a correct claim is fed into the masker. 
At this stage, we do not expect the masker to identify erroneous parts of the input claim. 
As shown in Figure \ref{fig.generated_claim}, the random strategy can produce much more diverse masked claims than the heuristic strategy, thus generating more diverse false claims. That is why the random strategy is more beneficial during testing.

\paragraph{Will different levels of masking granularity affect the performance of LIFE?} 
When using the T5 tokenizer, the input claim will be transformed into a subword sequence. 
Consequently, a complete word might be split into multiple subwords. 
For instance, if the text to be masked is `pleasingly large', after tokenization it will be split into `\textbf{\_}pleasing', `ly', `\textbf{\_}large'. 
If the masker's minimum masking granularity is at the subword level, we will use `\# \# \#' to represent the masked subwords. On the other hand, if the minimum masking granularity is at the word level, we will use `\# \#' to represent the masked words. 
Furthermore, we experiment with merging consecutive mask tokens. This means that each span is replaced with a single \# token, thereby `pleasingly large' will be represented as `\#'. 
Table \ref{tab.mask_span} shows that LIFE consistently performs well across different masking granularity, highlighting the robustness of our approach. 
In other experiments, we use the word level masking granularity and do not merge adjacent masked words.

\begin{table}[t] 
  \centering
 % \footnotesize
  \scriptsize
   \begin{tabular}{
    m{0.08\textwidth}<{\raggedright}
    m{0.05\textwidth}<{\raggedright}
    m{0.03\textwidth}<{\centering}
    m{0.04\textwidth}<{\centering}
    m{0.03\textwidth}<{\centering}
    m{0.03\textwidth}<{\centering}
    m{0.03\textwidth}<{\centering}
    }
    % \toprule
    \hline
     \multicolumn{2}{c}{\textbf{Mask Span}} & \multicolumn{4}{c} {\textbf{SARI Score}} & \multirow{2}{*}{\textbf{RG-2}} \\
    \cmidrule(lr){3-6} 
    \cmidrule(lr){1-2} 
   \textbf{Granularity}&  \textbf{Merge} & \textbf{Keep} &  \textbf{Delete} & \textbf{Add} & \textbf{Final}  & \\
   % \midrule
   \hline
   Subword &  & 74.37 & 93.10 & 28.07 & 65.18 & \textbf{66.75} \\
   Subword & \checkmark & 73.59 & 92.62 & 27.90 & 64.70 & 66.27\\
   % \midrule
   \hline
   Word &  & 75.23 & 91.88 & 29.67 & \textbf{65.59} & 66.51 \\
   Word & \checkmark & 73.94 & 93.05 & 27.48 & 64.82 & 66.30\\
    % \bottomrule
    \hline
 \end{tabular}
 \caption{Impact of masking granularities on LIFE. 
 }\label{tab.mask_span} 
\end{table}

\begin{table}[t] 
  \centering
 % \footnotesize
  \scriptsize
   \begin{tabular}{
    m{0.13\textwidth}<{\raggedright}
    m{0.03\textwidth}<{\centering}
    m{0.03\textwidth}<{\centering}
    m{0.03\textwidth}<{\centering}
    m{0.04\textwidth}<{\centering}
    m{0.04\textwidth}<{\centering}
    }
    % \toprule
    \hline
      \multirow{2}{*}{\textbf{\#} \textbf{Variants}} & \multicolumn{4}{c} {\textbf{SARI Score}} & \multirow{2}{*}{\textbf{RG-2}} \\
    \cmidrule(lr){2-5} 
    & \textbf{Keep} &  \textbf{Delete} & \textbf{Add} & \textbf{Final}  & \\
    % \midrule
    \hline
    1 LIFE &  75.23 & 91.88 & 29.67 & \textbf{65.59} & \textbf{66.51} \\
    \midrule
    2\quad -- LF & 73.65 & 93.98 & 26.65 & 64.76 & 66.38 \\
    3\quad -- FVCF & 74.80 & 91.63 & 28.78 & 65.07 & 66.10 \\
    4\quad -- LF \& FVCF  & 72.90 & 93.99 & 25.47 & 64.12 & 66.05 \\
    % \bottomrule
    \hline
 \end{tabular}
 \caption{Ablation study of LIFE on the test set. 
 }\label{tab.ablation} 
\end{table} 

\paragraph{Impact of Filters.}
We perform an ablation study to demonstrate the 
importance of filters. 
We first train the full model on the synthetic data that has been processed by the filters: LF and FVCF. 
The results, as shown in Table \ref{tab.ablation}, highlight two key points when compared to the full model (row 1):
(1) Eliminating LF (row 2) or FVCF (row 3) leads to performance drop.
(2) The removal of both filters results in the poorest performance among all variants.
These observations verify the effectiveness of the proposed filters. 
% The combined absence of both filters yields the poorest results across all variations.
% These observations serve to validate the efficacy of the proposed filters in our approach.

\subsection{Human Evaluation}
Apart from automatic evaluation, we also conduct a human evaluation to compare LIFE with the fully supervised T5-base, 8-shot T5-base, and 8-shot ChatGPT models. 
The fully supervised T5-base model employs gold evidence to rectify false claims, while the others use retrieved evidence. 
We randomly sample 50 samples from the test set and shuffle them to avoid bias. 
Following \citet{chen2023converge}, three annotators\footnote{All annotators are Ph.D. holders unrelated to our research.
} are asked to assess the revised claims using the following Boolean criteria: 
(1) Is it \textit{grammatically} correct? 
(2) Is it \textit{supported} by evidence? 
(3) Are the factual errors \textit{corrected}?
The final question, measuring the correction of factual errors, is the primary metric in our human evaluation. 
Table \ref{tab.human} shows that our proposed model outperforms the few-shot baselines on the corrected metric.
However, there still exists a gap between our model and the fully supervised baseline. 
Inter-annotator agreement measured by Fleiss' \textit{kappa} \cite{Fleiss1971MeasuringNS} is 0.859, implying almost perfect agreement ($>0.8$) \cite{Landis1977TheMO}.

\begin{table}[t] 
  \centering
 % \footnotesize
  \scriptsize
   \begin{tabular}{
    m{0.19\textwidth}<{\raggedright}
    m{0.07\textwidth}<{\raggedright}
    m{0.06\textwidth}<{\raggedright}
    m{0.05\textwidth}<{\raggedright}
    }
    % \toprule
    \hline
   \textbf{Models} & \textbf{Grammatical} &  \textbf{Supported} & \textbf{Corrected} \\
    % \midrule
    \hline
    \multicolumn{4}{l}{\textbf{Fully supervised models with gold evidence}}\\
    \quad T5-base & 100 & 89.3$^{\ast}$ & \underline{86.7}$^{\ast}$ \\
    % \midrule
    \hline
    \multicolumn{4}{l}{\textbf{ Models with retrieved evidence}}\\
    \quad 8-shot T5-base & 83.3$^{\ast}$ & 22.0$^{\ast}$ & 5.3$^{\ast}$ \\
    \quad 8-shot ChatGPT & 92.0$^{\ast}$ & 90.0$^{\ast}$ & 42.0$^{\ast}$ \\
    % \quad PivotFEC (ChatGPT) & 99.3 & 65.3 & \textbf{54.7} \\
    \quad LIFE  & 99.3 & 68.7 & \textbf{58.7} \\
    % \bottomrule
    \hline
 \end{tabular}
 \caption{ Human evaluation results (\%) on the test set.
 \textit{$p$}-values between results of our proposed model LIFE and the results marked with $^{\ast}$ are less than 0.01, indicating the differences between them are statistically significant. 
 % for the grammatical, supported and corrected scores. 
 }\label{tab.human} 
\end{table} 

\subsection{Case Study}
\begin{figure}
  \centering
    \includegraphics[width=0.46\textwidth]{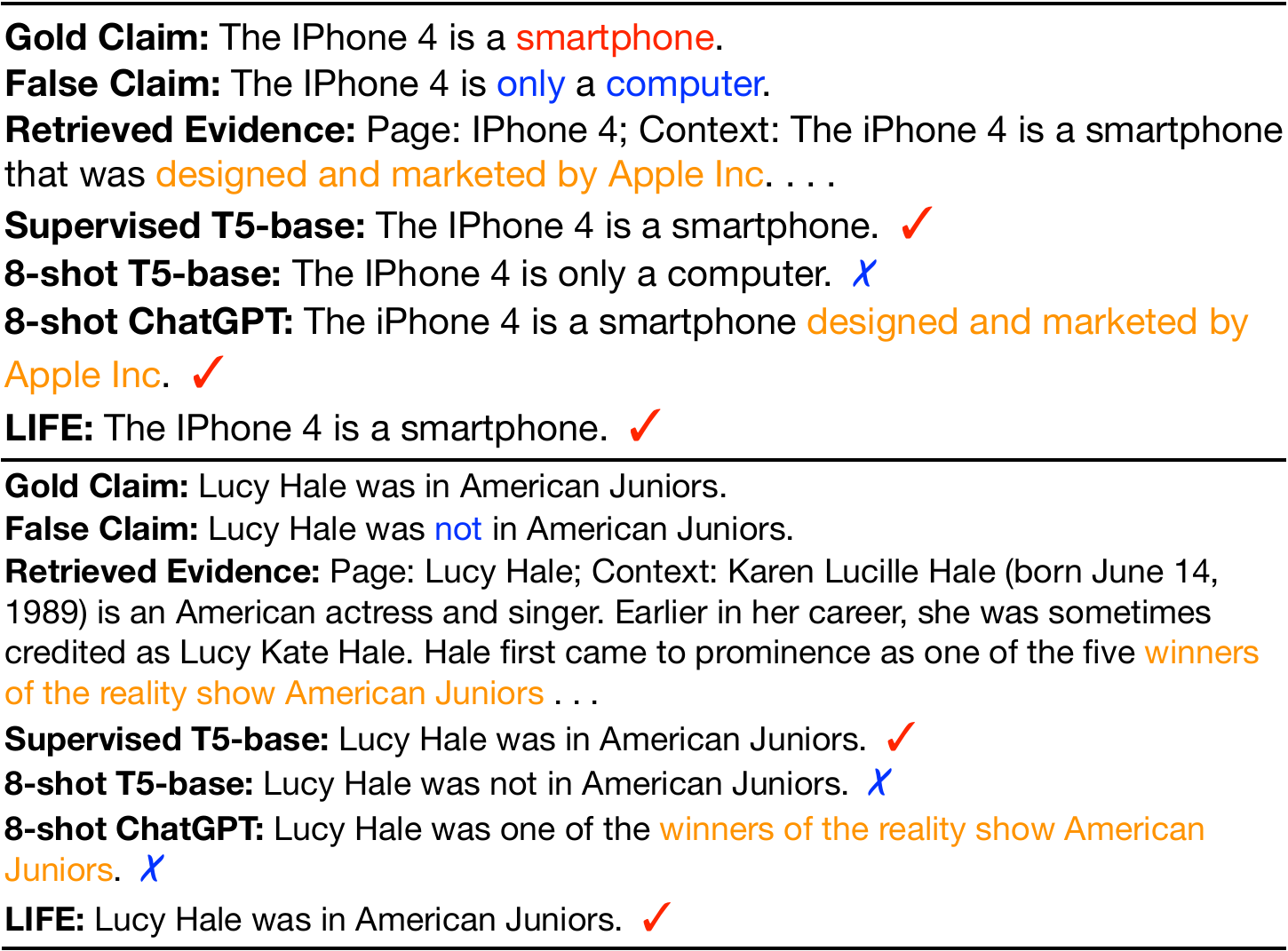} 
    \caption{ 
    Revised claims generated by our model and baselines  for false claims from the test set. 
    The supervised T5-base method corrects with gold  evidence, while others use retrieved evidence. We omit the gold evidence for simplicity 
    (blue text=factual errors; red text=correct modifications; orange text=content copied from the evidence).
      }\label{fig.case_study}
\end{figure}

In this subsection, we analyze the claims revised by our approach and baselines. 
As shown in Figure \ref{fig.case_study}, the supervised T5-base can correctly revise the factual errors in false claims, which represents the ceiling performance. However, the 8-shot T5-base method does not make any modifications, thus performing worst. 
While 8-shot ChatGPT can occasionally rectify factual errors within false claims, it typically involves directly copying content from the evidence into revised claims rather than making direct corrections to the erroneous parts. This tendency results in revised claims containing extra information. For instance, in the first example,  ``\textit{designed and marketed by Apple Inc}" is irrelevant to the correction. 
This phenomenon also explains why this method achieves a significant supported score in human evaluation, yet a lower corrected score.
In contrast, our approach can pinpoint factual errors and make precise corrections.

\section{Related Work}
% \subsection{Grammatical Error Correction} 
\textbf{Grammatical Error Correction} \cite{ng-etal-2014-conll, yuan-briscoe-2016-grammatical, bryant-etal-2017-automatic, awasthi-etal-2019-parallel, liu-etal-2021-neural} is meant to identify and rectify grammatical errors in written text, which is critical for aiding non-native speakers in enhancing their writing skills.  
% supporting language learners in refining their grammatical precision, and assisting professional writers in crafting error-free and polished content. 
Compared with grammatical error correction, factual error correction 
% \cite{shah2020automatic, thorne-vlachos-2021-evidence, chen2023converge}
aims to correct factual errors, such as incorrect dates, names, or historical events,  instead of grammatical errors. 
At the same time, \citet{he-etal-2023-pivotfec} proposed using LLMs, such as ChatGPT, as annotators \cite{he2023annollm} to inject factual errors into correct text to create synthetic data. In contrast, we suggest utilizing a trained corruptor to inject factual errors into the correct text, without relying on LLMs.
% \paragraph{Factual Error Correction} (FEC) \cite{shah2020automatic, thorne-vlachos-2021-evidence, chen2023converge} is specifically designed to  correct factual errors instead of grammatical errors. Factual errors includes incorrect dates, names, or historical events. 
% There are some works related to FEC, but rely on different underlying assumptions. Therefore, we cannot directly compare our work with theirs. 
% For example, \citet{huang-etal-2023-zero} assume the availability of gold standard evidence, and use additional QA datasets. 
% On the other hand, \citet{gao-etal-2023-rarr, chen2023purr} assume the perfection of retrieved evidence, and evaluate revised claims by checking whether they are supported by the retrieved evidence. 
% Following \citet{thorne-vlachos-2021-evidence}, we acknowledge the inadequacy and limitations of retrieved evidence, and evaluate revised claims by comparing them against gold claims (i.e., human-revised claims). 
\\
% \subsection{Retrieval-Augmented Generation} 
\textbf{Retrieval-augmented Generation} (RAG) \cite{10.5555/3495724.3496517} integrates information retrieval and language generation techniques to enhance the quality of generated content. 
\citet{he2022metric} use dense retrievers to fetch relevant sentences from an external corpus using specific keywords, improving lexically constrained text generation \cite{he2021xlentmcmc,he2021parallel}. 
% In this paradigm, one can utilize either sparse retrievers or dense retrievers to extract relevant information from a predefined database or a large corpus of text.
By incorporating external knowledge, RAG effectively mitigates the risk of generating inaccurate or nonsensical content. 
Factual error correction revises factual errors based on the retrieved evidence, thereby falling under the category of RAG.\\
% \subsection{Fact Verification} 
\textbf{Fact Verification}, also referred to as fact-checking, seeks to check whether a claim is supported or refuted by the given evidence, which has undergone extensive research in recent years. 
Researchers evaluate claims by analyzing both unstructured sources \cite{vlachos-riedel-2014-fact, wang-2017-liar, thorne-etal-2018-fever, wadden-etal-2020-fact} and structured sources \cite{chentabfact, iso-etal-2020-fact}. 
In contrast, factual error correction is  more challenging than fact verification, since it necessitates models not only to evaluate whether input claims contain factual errors, but also to pinpoint and rectify them. 

% Compared with fact verification, factual error correction is more challenging, since it not only requires models to assess whether the input claim contains factual errors, 
% but also requires them to identify factual errors and rectify them. 
\section{Conclusion and Future Work}
In this paper, we present LIFE, which improves FEC by learning to inject factual errors into correct claims. 
LIFE is a distantly supervised model comprising three components: masker, corruptor and corrector. 
The core of our approach involves training a corruptor to inject factual errors into correct claims using the `\textit{mask-then-corrupt}' strategy. This approach enables us to generate a substantial amount of synthetic data, which can be used to train the corrector. 
Our approach circumvents the bottleneck of identifying factual errors before making modifications encountered by previous distantly supervised models. 
Consequently, our proposed model, LIFE, significantly outperforms previous distantly supervised baselines and the few-shot LLMs, achieving a new SOTA result on FECDATA. 

However, the training of the corruptor depends on whether the masker can correctly identify factual errors. 
To be concrete, during training, if the masker fails to recognize any factual errors within the false claim, the corruptor simply needs to restore the masked words in the masked claim without requiring to inject additional factual errors. This is why the corruptor may fail to introduce factual errors into correct claims during testing. Therefore, researching how to accurately identify factual errors in false claims can be considered as a future research direction.

\section{Ethics Statement}
We take ethical considerations very seriously and strictly adhere to AAAI's ethics policy. This work concentrates on enhancing factual error correction, analyzed through publicly available datasets and evaluation methods. We guarantee the authenticity of our experimental results and the objectivity of our empirical conclusions.

\section{Acknowledgments}
This work is supported by National Natural Science Foundation of China (Grant No. 62202023), HKU-SCF FinTech Academy, Shenzhen-Hong Kong-Macao Science and Technology Plan Project (Category C Project: SGDX20210823103537030), and Theme-based Research Scheme of RGC, Hong Kong (T35-710/20-R). 
We would like to thank the anonymous reviewers for their constructive and informative feedback on this work.

% \bigskip
% \noindent Thank you for reading these instructions carefully. We look forward to receiving your electronic files!

\bibliography{aaai24}

\begin{thebibliography}{43}
\providecommand{\natexlab}[1]{#1}

\bibitem[{Awasthi et~al.(2019)Awasthi, Sarawagi, Goyal, Ghosh, and Piratla}]{awasthi-etal-2019-parallel}
Awasthi, A.; Sarawagi, S.; Goyal, R.; Ghosh, S.; and Piratla, V. 2019.
\newblock Parallel Iterative Edit Models for Local Sequence Transduction.
\newblock In \emph{Proceedings of EMNLP}, 4260--4270. Hong Kong, China: Association for Computational Linguistics.

\bibitem[{Brown et~al.(2020)Brown, Mann, Ryder, Subbiah, Kaplan, Dhariwal, Neelakantan, Shyam, Sastry, Askell, Agarwal, Herbert-Voss, Krueger, Henighan, Child, Ramesh, Ziegler, Wu, Winter, Hesse, Chen, Sigler, Litwin, Gray, Chess, Clark, Berner, McCandlish, Radford, Sutskever, and Amodei}]{NEURIPS2020_1457c0d6}
Brown, T.; Mann, B.; Ryder, N.; Subbiah, M.; Kaplan, J.~D.; Dhariwal, P.; Neelakantan, A.; Shyam, P.; Sastry, G.; Askell, A.; Agarwal, S.; Herbert-Voss, A.; Krueger, G.; Henighan, T.; Child, R.; Ramesh, A.; Ziegler, D.; Wu, J.; Winter, C.; Hesse, C.; Chen, M.; Sigler, E.; Litwin, M.; Gray, S.; Chess, B.; Clark, J.; Berner, C.; McCandlish, S.; Radford, A.; Sutskever, I.; and Amodei, D. 2020.
\newblock Language Models are Few-Shot Learners.
\newblock In Larochelle, H.; Ranzato, M.; Hadsell, R.; Balcan, M.; and Lin, H., eds., \emph{NIPS}, volume~33, 1877--1901. Curran Associates, Inc.

\bibitem[{Bryant, Felice, and Briscoe(2017)}]{bryant-etal-2017-automatic}
Bryant, C.; Felice, M.; and Briscoe, T. 2017.
\newblock Automatic Annotation and Evaluation of Error Types for Grammatical Error Correction.
\newblock In \emph{Proceedings of ACL}, 793--805. Vancouver, Canada: Association for Computational Linguistics.

\bibitem[{Cao et~al.(2021)Cao, Izacard, Riedel, and Petroni}]{cao2021autoregressive}
Cao, N.~D.; Izacard, G.; Riedel, S.; and Petroni, F. 2021.
\newblock Autoregressive Entity Retrieval.
\newblock In \emph{ICLR}.

\bibitem[{Chen et~al.(2023)Chen, Xu, Zeng, Sun, Li, and Xiao}]{chen2023converge}
Chen, J.; Xu, R.; Zeng, W.; Sun, C.; Li, L.; and Xiao, Y. 2023.
\newblock Converge to the Truth: Factual Error Correction via Iterative Constrained Editing.
\newblock In \emph{Proceedings of AAAI}.

\bibitem[{Chen et~al.(2017)Chen, Zhu, Ling, Wei, Jiang, and Inkpen}]{chen-etal-2017-enhanced}
Chen, Q.; Zhu, X.; Ling, Z.-H.; Wei, S.; Jiang, H.; and Inkpen, D. 2017.
\newblock Enhanced {LSTM} for Natural Language Inference.
\newblock In \emph{Proceedings of ACL}, 1657--1668. Vancouver, Canada: Association for Computational Linguistics.

\bibitem[{Chen et~al.(2020)Chen, Wang, Chen, Zhang, Wang, Li, Zhou, and Wang}]{chentabfact}
Chen, W.; Wang, H.; Chen, J.; Zhang, Y.; Wang, H.; Li, S.; Zhou, X.; and Wang, W.~Y. 2020.
\newblock TabFact: A Large-scale Dataset for Table-based Fact Verification.
\newblock In \emph{ICLR}.

\bibitem[{Chowdhery et~al.(2022)Chowdhery, Narang, Devlin, Bosma, Mishra, Roberts, Barham, Chung, Sutton, Gehrmann et~al.}]{chowdhery2022palm}
Chowdhery, A.; Narang, S.; Devlin, J.; Bosma, M.; Mishra, G.; Roberts, A.; Barham, P.; Chung, H.~W.; Sutton, C.; Gehrmann, S.; et~al. 2022.
\newblock PaLM: Scaling Language Modeling with Pathways.
\newblock \emph{arXiv preprint arXiv:2204.02311}.

\bibitem[{Devlin et~al.(2019)Devlin, Chang, Lee, and Toutanova}]{devlin-etal-2019-bert}
Devlin, J.; Chang, M.-W.; Lee, K.; and Toutanova, K. 2019.
\newblock {BERT}: Pre-training of Deep Bidirectional Transformers for Language Understanding.
\newblock In \emph{Proceedings of NAACL}, 4171--4186. Minneapolis, Minnesota: Association for Computational Linguistics.

\bibitem[{Fleiss(1971)}]{Fleiss1971MeasuringNS}
Fleiss, J.~L. 1971.
\newblock Measuring nominal scale agreement among many raters.
\newblock \emph{Psychological Bulletin}, 76(5): 378–382.

\bibitem[{He(2021)}]{he2021parallel}
He, X. 2021.
\newblock Parallel Refinements for Lexically Constrained Text Generation with {BART}.
\newblock In \emph{Proceedings of EMNLP}, 8653--8666. Online and Punta Cana, Dominican Republic: Association for Computational Linguistics.

\bibitem[{He et~al.(2022)He, Gong, Jin, Qi, Zhang, Jiao, Zhou, Cheng, Yiu, and Duan}]{he2022metric}
He, X.; Gong, Y.; Jin, A.-L.; Qi, W.; Zhang, H.; Jiao, J.; Zhou, B.; Cheng, B.; Yiu, S.; and Duan, N. 2022.
\newblock Metric-guided Distillation: Distilling Knowledge from the Metric to Ranker and Retriever for Generative Commonsense Reasoning.
\newblock In \emph{Proceedings of EMNLP}, 839--852. Abu Dhabi, United Arab Emirates: Association for Computational Linguistics.

\bibitem[{He et~al.(2023{\natexlab{a}})He, Jin, Ma, Yuan, and Yiu}]{he-etal-2023-pivotfec}
He, X.; Jin, A.-L.; Ma, J.; Yuan, Y.; and Yiu, S.~M. 2023{\natexlab{a}}.
\newblock {P}ivot{FEC}: Enhancing Few-shot Factual Error Correction with a Pivot Task Approach using Large Language Models.
\newblock In Bouamor, H.; Pino, J.; and Bali, K., eds., \emph{Findings of EMNLP}, 9960--9976. Singapore: Association for Computational Linguistics.

\bibitem[{He and Li(2021)}]{he2021xlentmcmc}
He, X.; and Li, V.~O. 2021.
\newblock Show Me How To Revise: Improving Lexically Constrained Sentence Generation with {XLNet}.
\newblock In \emph{Proceedings of AAAI}, volume~35, 12989--12997.

\bibitem[{He et~al.(2023{\natexlab{b}})He, Lin, Gong, Jin, Zhang, Lin, Jiao, Yiu, Duan, and Chen}]{he2023annollm}
He, X.; Lin, Z.; Gong, Y.; Jin, A.-L.; Zhang, H.; Lin, C.; Jiao, J.; Yiu, S.~M.; Duan, N.; and Chen, W. 2023{\natexlab{b}}.
\newblock AnnoLLM: Making Large Language Models to Be Better Crowdsourced Annotators.
\newblock \emph{arXiv preprint arXiv:2303.16854}.

\bibitem[{Iso, Qiao, and Li(2020)}]{iso-etal-2020-fact}
Iso, H.; Qiao, C.; and Li, H. 2020.
\newblock {F}act-based {T}ext {E}diting.
\newblock In \emph{Proceedings of ACL}, 171--182. Online: Association for Computational Linguistics.

\bibitem[{Karpukhin et~al.(2020)Karpukhin, Oguz, Min, Lewis, Wu, Edunov, Chen, and Yih}]{karpukhin-etal-2020-dense}
Karpukhin, V.; Oguz, B.; Min, S.; Lewis, P.; Wu, L.; Edunov, S.; Chen, D.; and Yih, W.-t. 2020.
\newblock Dense Passage Retrieval for Open-Domain Question Answering.
\newblock In \emph{Proceedings of EMNLP}, 6769--6781. Online: Association for Computational Linguistics.

\bibitem[{Landis and Koch(1977)}]{Landis1977TheMO}
Landis, J.~R.; and Koch, G.~G. 1977.
\newblock The measurement of observer agreement for categorical data.
\newblock \emph{Biometrics}, 33(1): 159--174.

\bibitem[{Levenshtein et~al.(1966)}]{levenshtein1966binary}
Levenshtein, V.~I.; et~al. 1966.
\newblock Binary codes capable of correcting deletions, insertions, and reversals.
\newblock In \emph{Soviet physics doklady}, volume~10, 707--710. Soviet Union.

\bibitem[{Lewis et~al.(2020)Lewis, Perez, Piktus, Petroni, Karpukhin, Goyal, K\"{u}ttler, Lewis, Yih, Rockt\"{a}schel, Riedel, and Kiela}]{10.5555/3495724.3496517}
Lewis, P.; Perez, E.; Piktus, A.; Petroni, F.; Karpukhin, V.; Goyal, N.; K\"{u}ttler, H.; Lewis, M.; Yih, W.-t.; Rockt\"{a}schel, T.; Riedel, S.; and Kiela, D. 2020.
\newblock Retrieval-Augmented Generation for Knowledge-Intensive NLP Tasks.
\newblock In \emph{Proceedings of NIPS}. Red Hook, NY, USA: Curran Associates Inc.
\newblock ISBN 9781713829546.

\bibitem[{Lin(2004)}]{lin-2004-rouge}
Lin, C.-Y. 2004.
\newblock {ROUGE}: A Package for Automatic Evaluation of Summaries.
\newblock In \emph{Text Summarization Branches Out}, 74--81. Barcelona, Spain: Association for Computational Linguistics.

\bibitem[{Liu et~al.(2019)Liu, Ott, Goyal, Du, Joshi, Chen, Levy, Lewis, Zettlemoyer, and Stoyanov}]{liu2019roberta}
Liu, Y.; Ott, M.; Goyal, N.; Du, J.; Joshi, M.; Chen, D.; Levy, O.; Lewis, M.; Zettlemoyer, L.; and Stoyanov, V. 2019.
\newblock RoBERTa: A Robustly Optimized BERT Pretraining Approach.
\newblock \emph{arXiv preprint arXiv:1907.11692}.

\bibitem[{Liu et~al.(2021)Liu, Yi, Sun, Yang, and Chua}]{liu-etal-2021-neural}
Liu, Z.; Yi, X.; Sun, M.; Yang, L.; and Chua, T.-S. 2021.
\newblock Neural Quality Estimation with Multiple Hypotheses for Grammatical Error Correction.
\newblock In Toutanova, K.; Rumshisky, A.; Zettlemoyer, L.; Hakkani-Tur, D.; Beltagy, I.; Bethard, S.; Cotterell, R.; Chakraborty, T.; and Zhou, Y., eds., \emph{Proceedings of NAACL}, 5441--5452. Online: Association for Computational Linguistics.

\bibitem[{Loshchilov and Hutter(2019)}]{loshchilovdecoupled}
Loshchilov, I.; and Hutter, F. 2019.
\newblock Decoupled Weight Decay Regularization.
\newblock In \emph{ICLR}.

\bibitem[{Maynez et~al.(2020)Maynez, Narayan, Bohnet, and McDonald}]{maynez-etal-2020-faithfulness}
Maynez, J.; Narayan, S.; Bohnet, B.; and McDonald, R. 2020.
\newblock On Faithfulness and Factuality in Abstractive Summarization.
\newblock In \emph{Proceedings of ACL}, 1906--1919. Online: Association for Computational Linguistics.

\bibitem[{Ng et~al.(2014)Ng, Wu, Briscoe, Hadiwinoto, Susanto, and Bryant}]{ng-etal-2014-conll}
Ng, H.~T.; Wu, S.~M.; Briscoe, T.; Hadiwinoto, C.; Susanto, R.~H.; and Bryant, C. 2014.
\newblock The {C}o{NLL}-2014 Shared Task on Grammatical Error Correction.
\newblock In \emph{Proceedings of the Eighteenth Conference on Computational Natural Language Learning: Shared Task}, 1--14. Baltimore, Maryland: Association for Computational Linguistics.

\bibitem[{Raffel et~al.(2020)Raffel, Shazeer, Roberts, Lee, Narang, Matena, Zhou, Li, and Liu}]{10.5555/3455716.3455856}
Raffel, C.; Shazeer, N.; Roberts, A.; Lee, K.; Narang, S.; Matena, M.; Zhou, Y.; Li, W.; and Liu, P.~J. 2020.
\newblock Exploring the Limits of Transfer Learning with a Unified Text-to-Text Transformer.
\newblock \emph{JMLR}, 21(1).

\bibitem[{Raunak, Menezes, and Junczys-Dowmunt(2021)}]{raunak-etal-2021-curious}
Raunak, V.; Menezes, A.; and Junczys-Dowmunt, M. 2021.
\newblock The Curious Case of Hallucinations in Neural Machine Translation.
\newblock In \emph{Proceedings of NAACL}, 1172--1183. Online: Association for Computational Linguistics.

\bibitem[{Ribeiro, Singh, and Guestrin(2016)}]{ribeiro-etal-2016-trust}
Ribeiro, M.; Singh, S.; and Guestrin, C. 2016.
\newblock {``}Why Should {I} Trust You?{''}: Explaining the Predictions of Any Classifier.
\newblock In \emph{Proceedings of NAACL}, 97--101. San Diego, California: Association for Computational Linguistics.

\bibitem[{See, Liu, and Manning(2017)}]{see-etal-2017-get}
See, A.; Liu, P.~J.; and Manning, C.~D. 2017.
\newblock Get To The Point: Summarization with Pointer-Generator Networks.
\newblock In \emph{Proceedings of ACL}, 1073--1083. Vancouver, Canada: Association for Computational Linguistics.

\bibitem[{Shah, Schuster, and Barzilay(2020)}]{shah2020automatic}
Shah, D.; Schuster, T.; and Barzilay, R. 2020.
\newblock Automatic Fact-guided Sentence Modification.
\newblock In \emph{Proceedings of AAAI}, volume~34, 8791--8798.

\bibitem[{Thorne and Vlachos(2021)}]{thorne-vlachos-2021-evidence}
Thorne, J.; and Vlachos, A. 2021.
\newblock Evidence-based Factual Error Correction.
\newblock In \emph{Proceedings of ACL}, 3298--3309. Online: Association for Computational Linguistics.

\bibitem[{Thorne et~al.(2018)Thorne, Vlachos, Christodoulopoulos, and Mittal}]{thorne-etal-2018-fever}
Thorne, J.; Vlachos, A.; Christodoulopoulos, C.; and Mittal, A. 2018.
\newblock {FEVER}: a Large-scale Dataset for Fact Extraction and {VER}ification.
\newblock In \emph{Proceedings of NAACL}, 809--819. New Orleans, Louisiana: Association for Computational Linguistics.

\bibitem[{Touvron et~al.(2023)Touvron, Lavril, Izacard, Martinet, Lachaux, Lacroix, Rozi{\`e}re, Goyal, Hambro, Azhar et~al.}]{touvron2023llama}
Touvron, H.; Lavril, T.; Izacard, G.; Martinet, X.; Lachaux, M.-A.; Lacroix, T.; Rozi{\`e}re, B.; Goyal, N.; Hambro, E.; Azhar, F.; et~al. 2023.
\newblock LLaMA: Open and Efficient Foundation Language Models.
\newblock \emph{arXiv preprint arXiv:2302.13971}.

\bibitem[{Vaswani et~al.(2017)Vaswani, Shazeer, Parmar, Uszkoreit, Jones, Gomez, Kaiser, and Polosukhin}]{transformer}
Vaswani, A.; Shazeer, N.; Parmar, N.; Uszkoreit, J.; Jones, L.; Gomez, A.~N.; Kaiser, L.~u.; and Polosukhin, I. 2017.
\newblock Attention is All you Need.
\newblock In Guyon, I.; Luxburg, U.~V.; Bengio, S.; Wallach, H.; Fergus, R.; Vishwanathan, S.; and Garnett, R., eds., \emph{NIPS}, volume~30. Curran Associates, Inc.

\bibitem[{Vlachos and Riedel(2014)}]{vlachos-riedel-2014-fact}
Vlachos, A.; and Riedel, S. 2014.
\newblock Fact Checking: Task definition and dataset construction.
\newblock In \emph{Proceedings of the {ACL} 2014 Workshop on Language Technologies and Computational Social Science}, 18--22. Baltimore, MD, USA: Association for Computational Linguistics.

\bibitem[{Wadden et~al.(2020)Wadden, Lin, Lo, Wang, van Zuylen, Cohan, and Hajishirzi}]{wadden-etal-2020-fact}
Wadden, D.; Lin, S.; Lo, K.; Wang, L.~L.; van Zuylen, M.; Cohan, A.; and Hajishirzi, H. 2020.
\newblock Fact or Fiction: Verifying Scientific Claims.
\newblock In \emph{Proceedings of EMNLP}, 7534--7550. Online: Association for Computational Linguistics.

\bibitem[{Wang(2017)}]{wang-2017-liar}
Wang, W.~Y. 2017.
\newblock {``}Liar, Liar Pants on Fire{''}: A New Benchmark Dataset for Fake News Detection.
\newblock In \emph{Proceedings of ACL}, 422--426. Vancouver, Canada: Association for Computational Linguistics.

\bibitem[{Wei et~al.(2022{\natexlab{a}})Wei, Tay, Bommasani, Raffel, Zoph, Borgeaud, Yogatama, Bosma, Zhou, Metzler et~al.}]{wei2022emergent}
Wei, J.; Tay, Y.; Bommasani, R.; Raffel, C.; Zoph, B.; Borgeaud, S.; Yogatama, D.; Bosma, M.; Zhou, D.; Metzler, D.; et~al. 2022{\natexlab{a}}.
\newblock Emergent Abilities of Large Language Models.
\newblock \emph{arXiv preprint arXiv:2206.07682}.

\bibitem[{Wei et~al.(2022{\natexlab{b}})Wei, Wang, Schuurmans, Bosma, Chi, Le, and Zhou}]{wei2022chain}
Wei, J.; Wang, X.; Schuurmans, D.; Bosma, M.; Chi, E.; Le, Q.; and Zhou, D. 2022{\natexlab{b}}.
\newblock Chain-of-Thought Prompting Elicits Reasoning in Large Language Models.
\newblock In \emph{NIPS}.

\bibitem[{Wolf et~al.(2019)Wolf, Debut, Sanh, Chaumond, Delangue, Moi, Cistac, Rault, Louf, Funtowicz, and Brew}]{Wolf2019HuggingFacesTS}
Wolf, T.; Debut, L.; Sanh, V.; Chaumond, J.; Delangue, C.; Moi, A.; Cistac, P.; Rault, T.; Louf, R.; Funtowicz, M.; and Brew, J. 2019.
\newblock HuggingFace's Transformers: State-of-the-art Natural Language Processing.
\newblock \emph{arXiv preprint arXiv:1910.03771}.

\bibitem[{Xu et~al.(2016)Xu, Napoles, Pavlick, Chen, and Callison-Burch}]{xu-etal-2016-optimizing}
Xu, W.; Napoles, C.; Pavlick, E.; Chen, Q.; and Callison-Burch, C. 2016.
\newblock Optimizing Statistical Machine Translation for Text Simplification.
\newblock \emph{Transactions of the Association for Computational Linguistics}, 4: 401--415.

\bibitem[{Yuan and Briscoe(2016)}]{yuan-briscoe-2016-grammatical}
Yuan, Z.; and Briscoe, T. 2016.
\newblock Grammatical error correction using neural machine translation.
\newblock In \emph{Proceedings of NAACL}, 380--386. San Diego, California: Association for Computational Linguistics.

\end{thebibliography}

\clearpage
\appendix{\Large{\textbf{Appendix}}}
\section{Few-shot Prompts Used by LLMs for FEC}\label{sec.fec_prompt}

Table \ref{tab.fec.negate_few-shot} shows the few-shot exemplars prompt used by LLMs for the FEC task.

\begin{table*}[h] 
  \centering
 \footnotesize
  % \scriptsize
  % \tiny
   \begin{tabular}{
    m{0.96\textwidth}
    }
    \toprule
    \textbf{Evidence:} The Lion King; The story takes place in a kingdom of lions in Africa and was influenced by William Shakespeare 's Hamlet .\\
    The Lion King; The Lion King tells the story of Simba , a young lion who is to succeed his father , Mufasa , as King of the Pride Lands ; however , after Simba 's uncle Scar murders Mufasa , Simba is manipulated into thinking he was responsible and flees into exile .\\
    \textbf{Mutated claim:} The Lion King has nothing to do with lions.\\
    \textbf{Original claim:}  The Lion King is about lions.\\
    \\
    \textbf{Evidence:} Indiana Jones; Henry Walton `` Indiana '' Jones Jr. ( also shortened to Indy ) is a fictional character and the protagonist of the Indiana Jones franchise .\\
    Indiana Jones; George Lucas created the character in homage to the action heroes of 1930s film serials .\\
    \textbf{Mutated claim:} Indiana Jones is real.\\
    \textbf{Original claim:}  Indiana Jones is fictional.\\
    \\
    \textbf{Evidence:} Scott Eastwood; Scott Eastwood ( born Scott Clinton Reeves ; March 21 , 1986 ) is an American actor , model , and professional skydiver .\\
    Scott Eastwood; He has also been the model for the fragrance Cool Water by Davidoff .\\
    \textbf{Mutated claim:} Scott Eastwood was incapable of working as a model.\\
    \textbf{Original claim:}  Scott Eastwood worked as a model.\\
    \\
    \textbf{Evidence:} Akshay Kumar; Kumar is also a producer and martial artist who has appeared in over a hundred Hindi films .\\
    Akshay Kumar; Having done so , he has established himself as a leading contemporary actor of Hindi cinema .\\
    \textbf{Mutated claim:} Akshay Kumar does not work in Hindi cinema.\\
    \textbf{Original claim:}  Akshay Kumar works in Hindi cinema.\\
    \\
    \textbf{Evidence:} Gorillaz; Gorillaz are an English virtual band created in 1998 by musician Damon Albarn and artist Jamie Hewlett .\\
    Virtual band; In music , a virtual band ( also called a virtual group , cartoon group , or cartoon band ) is any group whose members are not corporeal musicians , but animated characters .\\
    \textbf{Mutated claim:} Gorillaz is a German live band.\\
    \textbf{Original claim:}  Gorillaz is a British virtual band.\\
    \\
    \textbf{Evidence:} Grant Gustin; Thomas Grant Gustin ( born January 14 , 1990 ) is an American actor , singer , and dancer .\\
    Grant Gustin; He is known for his role as Barry Allen / the Flash ( based on the DC Comics character of the same name ) on the CW series The Flash and Arrow , both in the Arrowverse television franchise , and his role as Sebastian Smythe on the Fox series Glee .\\
    \textbf{Mutated claim:} Grant Gustin is only a writer.\\
    \textbf{Original claim:}  Grant Gustin is a singer.\\
    \\
    \textbf{Evidence:} RB Leipzig; RasenBallsport Leipzig e.V. , commonly known as RB Leipzig , is a German association football club based in Leipzig , Saxony .\\
    Football in Germany; Football is the most popular sport in Germany .\\
    \textbf{Mutated claim:} RB Leipzig plays the least popular German sport.\\
    \textbf{Original claim:}  RB Leipzig plays the most popular German sport.\\
    \\
    \textbf{Evidence:} One World Trade Center; One World Trade Center ( also known as 1 World Trade Center , 1 WTC or Freedom Tower ) is the main building of the rebuilt World Trade Center complex in Lower Manhattan , New York City .\\
    World Trade Center (2001–present); The original World Trade Center featured the landmark Twin Towers , which opened in 1973 , and were the tallest buildings in the world at their completion .\\
    \textbf{Mutated claim:} One World Trade Center opened in 1876.\\
    \textbf{Original claim:}   One World Trade Center opened in 2014.\\
    \\
    \textbf{Evidence:} \{\textbf{evidence}\} \\
    \textbf{Mutated claim:} \{\textbf{mutated claim}\}  \\
    \textbf{Original claim:}  \\
    \bottomrule
 \end{tabular}
 \caption{Few-shot exemplars prompt used by LLMs for the FEC task.}\label{tab.fec.negate_few-shot} 
\end{table*}

\end{document}